\documentclass[runningheads]{llncs}
\usepackage[T1]{fontenc}
\usepackage{graphicx}
\usepackage{makecell}
\usepackage{subcaption}
\usepackage{diagbox}

\begin{document}
\title{Counteracting temporal attacks in {Video Copy Detection}}
%
%
\author{Katarzyna Fojcik\inst{1}\orcidID{0000-0002-6627-742X} \and
Piotr Syga\inst{1}\orcidID{0000-0002-0266-5802}}
\authorrunning{K. Fojcik \and P. Syga}
%
\institute{Department of Artificial Intelligence, Wroclaw University of Science and Technology, Wroclaw, Poland}
\maketitle              
\begin{abstract}
Video Copy Detection (VCD) plays a crucial role in copyright protection and content verification by identifying duplicates and near-duplicates in large-scale video databases. The META AI Challenge on video copy detection provided a benchmark for evaluating state-of-the-art methods, with the Dual-level detection approach emerging as a winning solution. This method integrates Video Editing Detection and Frame Scene Detection to handle adversarial transformations and large datasets efficiently. However, our analysis reveals significant limitations in the VED component, particularly in its ability to handle exact copies. Moreover, Dual-level detection shows vulnerability to temporal attacks. To address it, we propose an improved frame selection strategy based on local maxima of interframe differences, which enhances robustness against adversarial temporal modifications while significantly reducing computational overhead. Our method achieves an increase of 1.4 to 5.8 times in efficiency over the standard 1 FPS approach. Compared to Dual-level detection method, our approach maintains comparable micro-average precision ($\mu$AP) while also demonstrating improved robustness against temporal attacks. Given 56\% reduced representation size and the inference time of more than 2 times faster, our approach is more suitable to real-world resource restriction.

\end{abstract}
\section{Introduction}
Video content has become ubiquitous in the digital age, serving as a primary medium for entertainment, education, and communication. As video-sharing platforms grow in popularity, the need for effective copyright protection mechanisms becomes increasingly important. Video copy detection systems aim to identify duplicates or near-duplicates of videos, enabling rights holders to enforce intellectual property laws and combat unauthorized usage. Aside from copyright protection, copy detection is an important tool in reducing database redundancies or localizing the original recordings of DeepFake attacks. 

The dual-level detection method, as described in~\cite{pizzi20242023,wang2023dual} was recognized as the winning solution in the prestigious META AI Challenge on video copy detection~\cite{pizzi20242023}. This approach combines Video Editing Detection (VED) with Frame Scene Detection (FSD) to address the challenges posed by large-scale datasets and adversarial transformations. Specifically, VED is designed to differentiate unedited videos by assigning them descriptors with small norms and negative bias terms instead of deriving descriptors using a trained model. This significantly reduces the computation time, but the assumption that an unedited video definitely does not contain copied fragments from a database of reference videos is naive. 
For edited videos, FSD captures situation when multiple
scenes are concatenated along edge and split the scenes in
one frame.

The dataset used in the original study is representative of real-world scenarios and contains edited and unedited videos. It includes a diverse range of transformations, such as scaling, cropping, and temporal shifts, as well as adversarial attacks such as frame duplication and deletion. These variations provide a robust benchmark for evaluating video copy detection methods. Despite challenge winners' method strong performance on this dataset, our experiments reveal critical limitations of the dual-level detection method, particularly in its VED component and frame extraction approach.

In this article, we present a detailed analysis of the shortcomings of the original method and propose enhancements to improve its efficiency and robustness. By leveraging local maxima from interframe differences for frame extraction and evaluating the system's resistance to temporal attacks, we aim to provide a more practical and resilient solution for video copy detection.

\section{Related work}
Video copy detection has been an active area of research for more than a decade, with early methods focusing on descriptors such as histograms and block-based matching techniques. As computational power and storage capabilities improved, researchers shifted towards feature-based approaches, utilizing local descriptors like Scale-Invariant Feature Transform (SIFT) and Speeded-Up Robust Features (SURF). These methods enabled more accurate matching of video fragments, even under transformations such as scaling, cropping, and rotation~\cite{wu2007practical,kordopatis2017near2}.

With the advent of deep learning, video copy detection has undergone a paradigm shift. Convolutional Neural Networks (CNNs) like VGG~\cite{simonyan2014very}, ResNet~\cite{he2016deep}, and models such as ViSiL~\cite{kordopatis2019visil} have significantly advanced frame-level feature extraction for video similarity learning. Additionally, architectures like 3D-CNNs~\cite{li2020two} and encoder–decoder ConvLSTM models~\cite{chiang2022multi} capture spatiotemporal dynamics, improving robustness to temporal edits. Advanced methods now incorporate transformer-based architectures, such as Video ViT~\cite{arnab2021vivit}, and self-supervised learning frameworks like 3D-CSL~\cite{deng20233d} to further enhance the system's ability to detect heavily edited and transformed videos~\cite{deng20233d,deng2024differentiable,black2023vader}. Lightweight approaches, including multi-teacher distillation frameworks~\cite{ma2024let} and compact Siamese neural networks~\cite{fojcik2025extremely}, have recently gained attention for achieving high efficiency and scalability on large datasets.

The importance of dynamic frame selection based on interframe differences to create highly compact video representations was shown in~\cite{fojcik2025extremely}. This approach reduces computational complexity while preserving the core structure of the video, enabling efficient storage and retrieval even under severe distortions such as extreme compression or resizing. Such strategies demonstrate the potential for real-time applications, where memory and computational constraints are critical. We extend this approach to show robustness against temporal attacks.

On the other hand,~\cite{ZHONG2024103863} proposes a CNN-LSTM hybrid model for human action recognition, evaluated on UCF101 and HMDB51 datasets. Their method improves accuracy by up to 3.5\% over CNN-only baselines but suffers from high computational complexity and sensitivity to viewpoint variations, making the solution unusable in real-world applications.
The authors of~\cite{kim2024} also indicate time and memory limitations and propose Relational Self-supervised Distillation with Compact Descriptors (RDCD) for image copy detection, using a lightweight network and compact descriptors to improve efficiency. Their method employs relational self-supervised distillation to transfer knowledge from a large teacher network (ResNet-50) to a smaller student network (EfficientNet-B0) and introduces contrastive learning with a hard negative loss to mitigate dimensional collapse. Evaluated on the DISC2021, Copydays (CD10K), and NDEC datasets, RDCD improves micro average precision (µAP) by about 5.0\%, depending on the descriptor sizes over the baseline while maintaining competitive performance with significantly smaller descriptor sizes. However, the reliance on knowledge distillation from a large network and additional computational costs during training remain limitations.
Despite these advances, challenges persist. Many systems struggle with efficiency when processing large-scale datasets, as feature extraction and matching remain computationally expensive. For example, as demonstrated in~\cite{fojcik2025extremely}, widely used models like ViSiL generate descriptors of 2025kB for a typical 30-second video, with an inference speed of approximately 6.5 samples per second (sps), where one sample corresponds to a 30-second video. In contrast, the method proposed in~\cite{fojcik2025extremely} achieves a descriptor size of just 1.875kB while delivering an inference speed of up to 178.6 sps with satisfactory accuracy. Similarly, the authors of~\cite{deng20233d} highlight that prevalent models such as VRL-F and TCA-F (both frame-based matching) require over 10 seconds to generate a single descriptor for a sample from the FIVR-200K dataset, with descriptor sizes spanning several megabytes. ViSiL, by comparison, is even less efficient, taking about 10 times longer and producing descriptors approximately 10 times larger.  

Robustness against adversarial transformations—such as frame manipulation, extreme compression, or rotation—also varies widely among methods. The dual-level detection method, incorporating Video Embedding Descriptors (VED) for unedited video identification and Fragment Similarity Detection (FSD) for edited content, addresses some of these issues. However, as highlighted in our analysis, these components require further refinement to ensure robustness and scalability in real-world scenarios.

\section{The META AI Challenge Dataset and Baseline Model}
The META AI Challenge on video copy detection, held in 2023, provided a rigorous platform for evaluating state-of-the-art methods in video similarity analysis. The challenge was designed to push the boundaries of video copy detection and localization, attracting top research teams from around the world. It included two tracks: video copy detection (VCD) and video copy localization (VCL). The VCD track focused on identifying whether two videos shared copied content, while the VCL track required participants to localize the temporal segments of shared content within video pairs. In this paper we focus on VCD track.

\begin{figure}[ht!]
    \centering 

\begin{subfigure}{0.48\textwidth}
  \includegraphics[width=\linewidth]{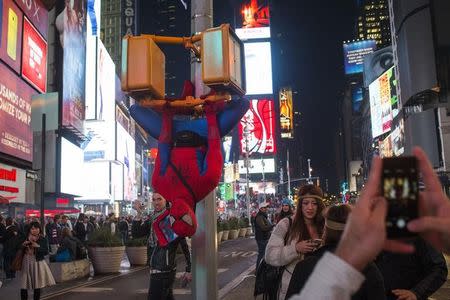}
  \caption{original}
  \label{original}
\end{subfigure}\hfil 
\begin{subfigure}{0.48\textwidth}
  \includegraphics[width=\linewidth]{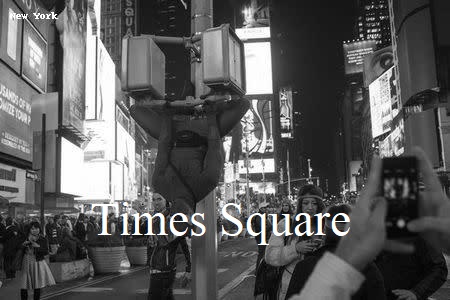}
  \caption{grayscale, overlay text}
  \label{gray_text}
\end{subfigure} 


\begin{subfigure}{0.48\textwidth}
  \includegraphics[width=\linewidth]{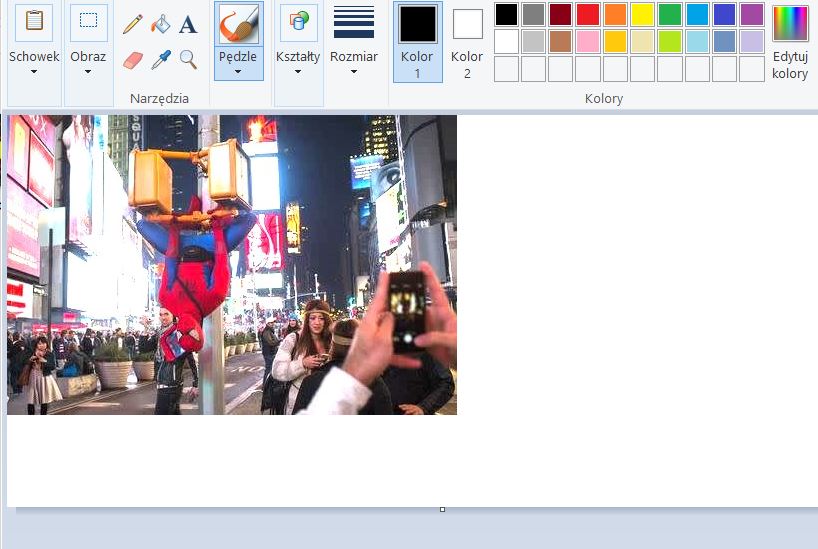}
  \caption{brightness, overlay onto screenshot}
  \label{bright_screenshot} 
\end{subfigure}\hfil
\begin{subfigure}{0.48\textwidth}
  \includegraphics[width=\linewidth]{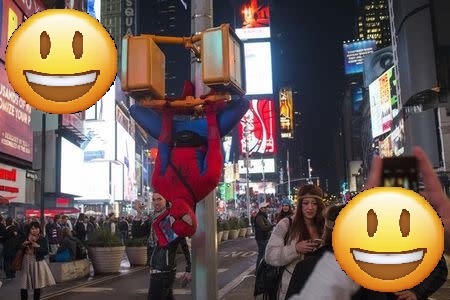}
  \caption{flip, overlay emoji}
  \label{muti_noise}
\end{subfigure} 


\begin{subfigure}{0.48\textwidth}
  \includegraphics[width=\linewidth]{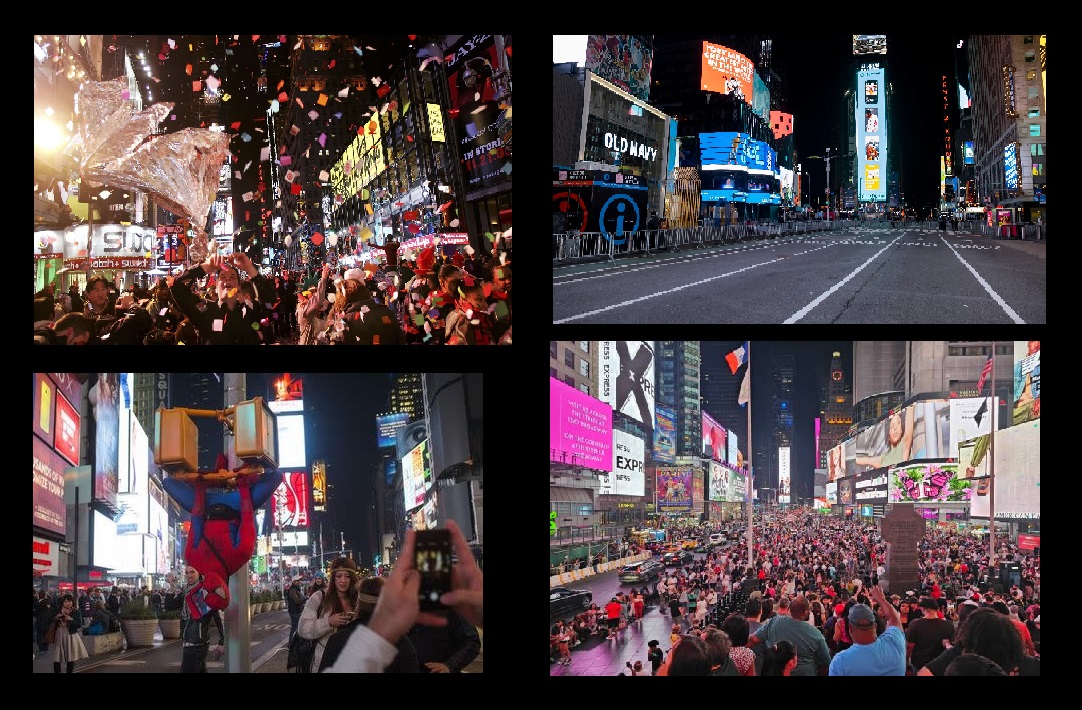}
  \caption{rooms}
  \label{rooms}
\end{subfigure}\hfil 
\begin{subfigure}{0.48\textwidth}
  \includegraphics[width=\linewidth]{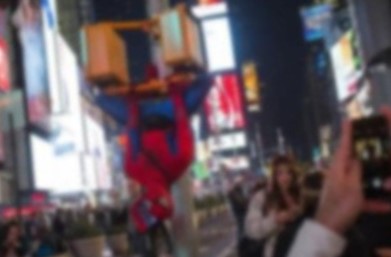}
  \caption{rotate, blur}
  \label{rotate_blur}
\end{subfigure}
\caption{Example of video frame with applied transformations.}
\label{fig:transformations}
\end{figure}

The participants were constrained by computational resource limits to ensure solutions were scalable to real-world scenarios. The competition emphasized practical applications, such as content moderation, copyright protection, and misinformation detection, highlighting the importance of efficient and accurate solutions in this domain. The challenge introduced a benchmark dataset-DVSC2023 and a strong baseline model to facilitate evaluation and comparison of methods. Although the official competition has ended, Meta AI Video Similarity Challenge and DVSC2023 are still available in the form of Open Arena\footnote{https://www.drivendata.org/competitions/219/competition-meta-vsc-desc-open/}\footnote{https://www.drivendata.org/competitions/220/competition-meta-vsc-match-open/}.

\subsection{Dataset}
The DVSC2023 dataset was created using videos from the YFCC100M collection, filtered to ensure Creative Commons licenses and to exclude videos that were too short or low-resolution. It contains a mixture of reference and query videos, with query videos being transformed versions of reference videos or distractors. The videos were modified using various augmentations, such as spatial updates (cropping, resizing), temporal edits (speed changes, frame alternation), and complex transformations like screen capture simulation. These transformations were applied to create challenging scenarios for detecting copies. Examples of video frames copied with applied transformations are presented in~Fig.~\ref{fig:transformations}. The data is partitioned into Training Split, that contains 8,404 query videos and 40,311 reference videos, with 2,708 queries containing copied segments, Validation Split, consisting of  8,295 query videos and 40,318 reference videos, with 2,641 queries containing copied segments, and Test Split of 8,015 query videos evaluated against 2,519 reference videos, with 1,840 queries containing copied segments.
Additionally, 6,475, 6,369, and 6,175 distractor queries are included in the training, validation, and test splits, respectively. These queries ensure that most queries contain no copied segments, replicating real-world conditions.

\noindent The contestants received only the training and validation splits before submitting their methods. Training data came with match results for all query-reference pairs, while validation match results were hidden and known only to the organizers. After submission, participants received feedback on validation split performance based on the ground truth. In Phase 2 of the challenge, the unseen test data was used for the final evaluation and leaderboard ranking, ensuring fair comparison across methods.
\subsection{Baseline Model}
The baseline model leverages the Self-Supervised Copy Detection (SSCD) descriptor~\cite{pizzi2022self}, which extracts frame descriptors at a rate of one frame per second after resizing and cropping. These descriptors undergo score normalization to mitigate false matches, enhancing reliability. The matching process is evaluated using the micro-average precision–recall ($\mu$AP) metric, which jointly assesses the ranked prediction lists of all queries based on their confidence scores. Precision and recall are calculated at each rank, with a positive when the query-reference pair corresponds to a ground-truth match involving copied segments.

\section{Limitations of the Original Method}
It is worth noting, that DVSC2023 dataset for the challenge does not include queries with exact duplicate fragments (unedited) of reference videos. However, it contains video pairs that are largely similar but exhibit subtle differences, distinguishing them as distinct videos rather than identical copies, e.g., videos may share the same background but feature different people.

The dual-level method uses VED to identify unedited videos, assigning random descriptors with small norms and negative bias terms to those that are unedited. Hence, it is able to handle these challenging video pairs, reporting a 5\% improvement in matching accuracy. However, when tested under real-world conditions, the method reveals a significant limitation. Our experiments show that VED consistently misclassifies exact video copies from the reference set as non-copies. We tested dual-level performance on 100 queries, which were exact copies of chosen 100 references video. None of them was recognized as a copy. This highlights a fundamental flaw in the current implementation of VED, making it unsuitable for practical applications.


Additionally, the authors deterministically extract one frame from each second of the video, which is a common approach used also in the baseline model of challenge organizers. Although straightforward, this approach is vulnerable to targeted temporal attacks, as discussed in Sect.~\ref{sect:results}.

\section{Proposed Method\label{sect:method}}
\subsection{Improved Frame Extraction}

To reduce computation and memory costs, we reduced the number of video frames while maintaining key data representation. Instead of the standard method of selecting one frame per second (fps), we focused on scene changes using the interframe difference curve. This was calculated as the sum of absolute pixel-wise differences between consecutive frames, averaged by the number of pixels.

\begin{figure}
\includegraphics[width=\textwidth]{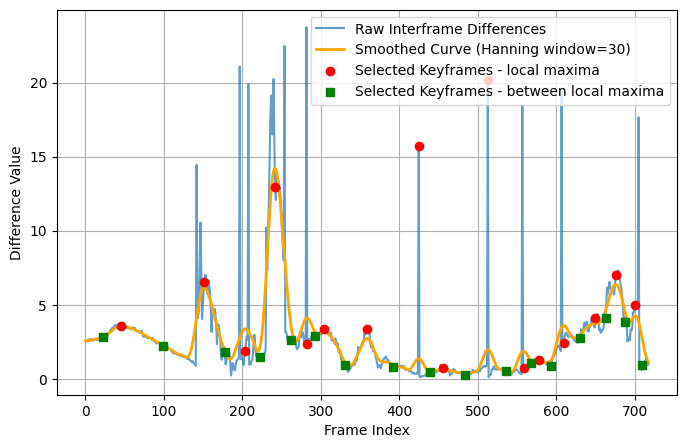}
\caption{Interframe differences curve before and after smoothing with Hanning window of size 30, and selected frames of a sample video.} \label{fig:window30}
\end{figure}

\begin{figure}
\includegraphics[width=\textwidth]{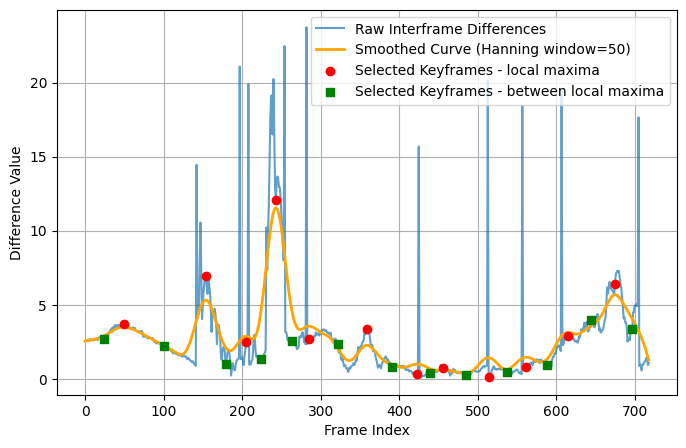}
\caption{Interframe differences curve before and after smoothing with Hanning window of size 50, and selected frames of a sample video.} \label{fig:window50}
\end{figure}

\begin{figure}
\includegraphics[width=\textwidth]{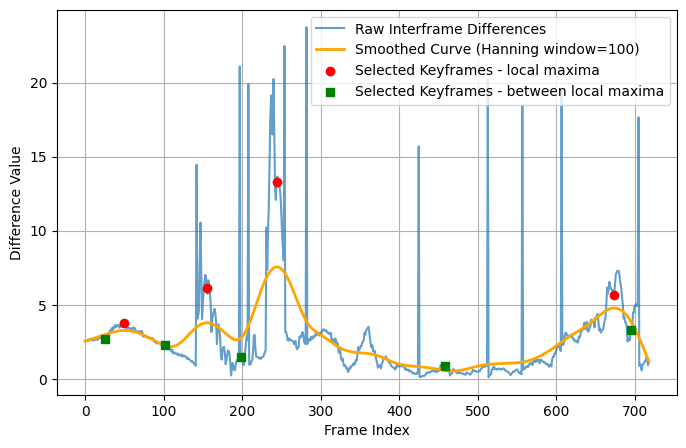}
\caption{Interframe differences curve before and after smoothing with Hanning window of size 100, and selected frames of a sample video.} \label{fig:window100}
\end{figure}

We tested two frame selection approaches. In first, we select frames which are local maxima of the interframe difference curve. The second approach used frames in the middle between two local maxima of the interframe difference curve. Both methods used Hanning window smoothing to reduce noise and highlight significant changes. Figs.~\ref{fig:window30},~\ref{fig:window50},~\ref{fig:window100} show the interframe differences curve before and after smoothing, as well as the selected frames for the first 30 seconds of an example video\footnote{https://www.youtube.com/watch?v=Os7s1HuNRyw}. The results and analysis of the influence of the window size are depicted in Tab.~\ref{tab1}.

\begin{figure}[ht!]
    \centering 

\begin{subfigure}{\textwidth}
  \includegraphics[height=3cm]{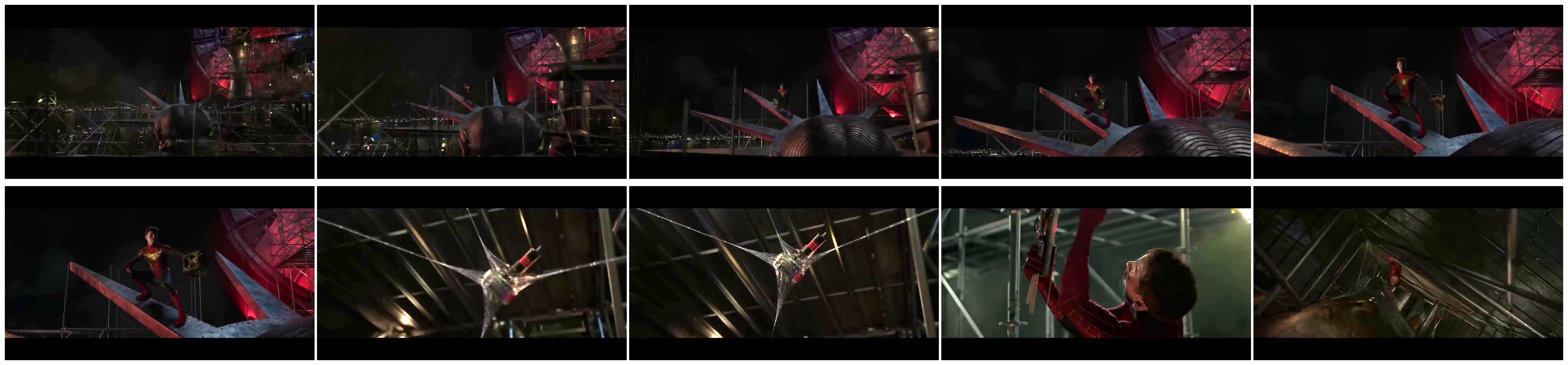}
  \caption{method: one frame per second}
  \label{original}
\end{subfigure}

\begin{subfigure}{\textwidth}
  \includegraphics[height=1.5cm]{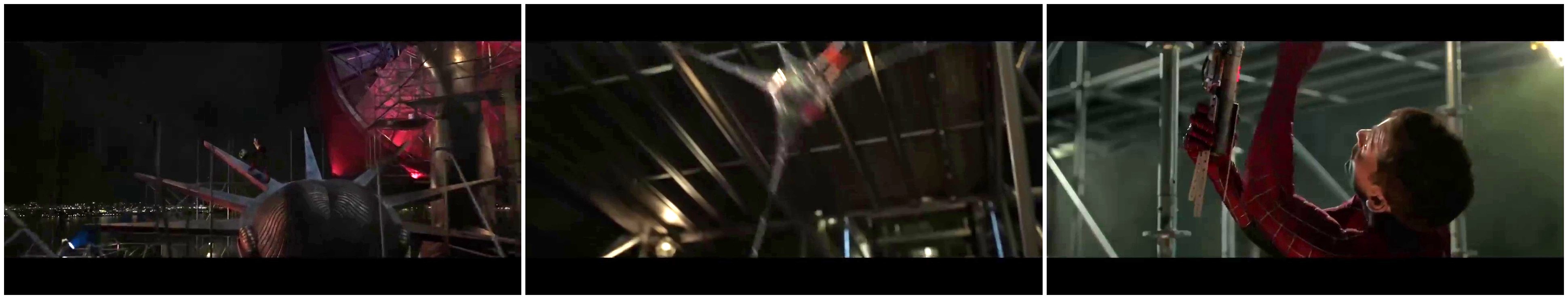}
  \caption{method: local maxima, hanning window: 30}
  \label{original}
\end{subfigure}
\begin{subfigure}{\textwidth}
  \includegraphics[height=1.5cm]{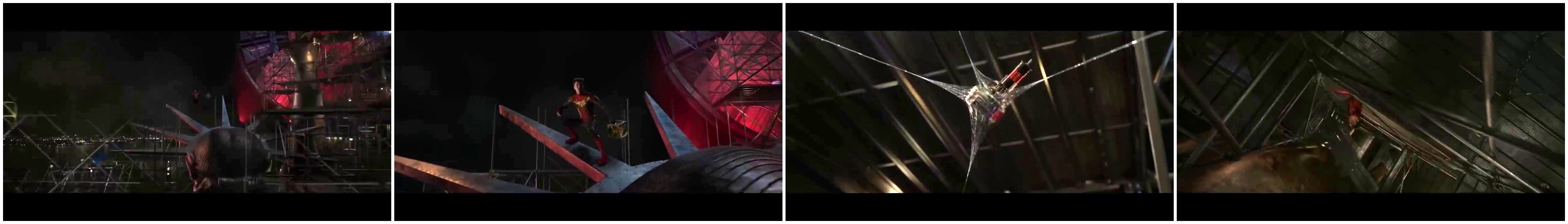}
  \caption{method: between local maxima, hanning window: 30}
  \label{gray_text}
\end{subfigure} 


\begin{subfigure}{\textwidth}
  \includegraphics[height=1.5cm]{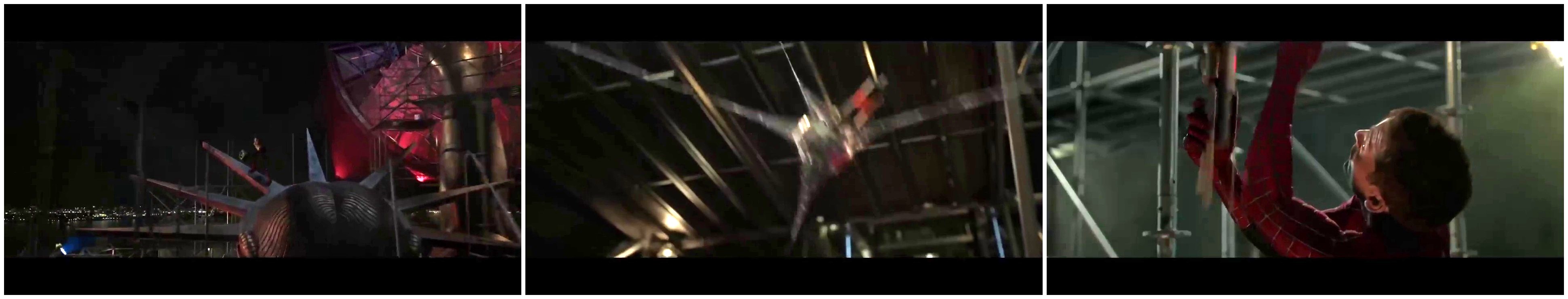}
  \caption{method: local maxima, hanning window: 50}
  \label{original}
\end{subfigure}
\begin{subfigure}{\textwidth}
  \includegraphics[height=1.5cm]{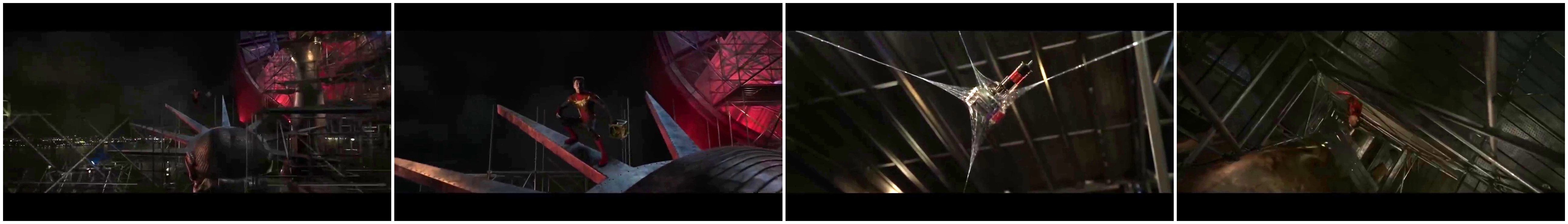}
  \caption{method: between local maxima, hanning window: 50}
  \label{gray_text}
\end{subfigure} 

\begin{subfigure}{\textwidth}
  \includegraphics[height=1.5cm]{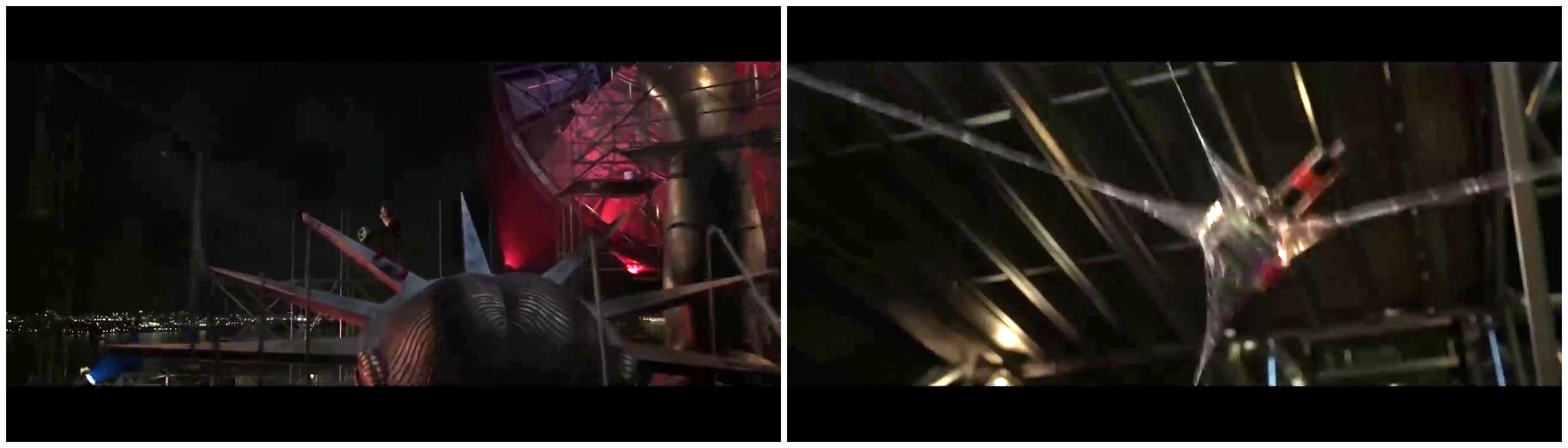}
  \caption{method: local maxima, hanning window: 100}
  \label{original}
\end{subfigure}
\begin{subfigure}{\textwidth}
  \includegraphics[height=1.5cm]{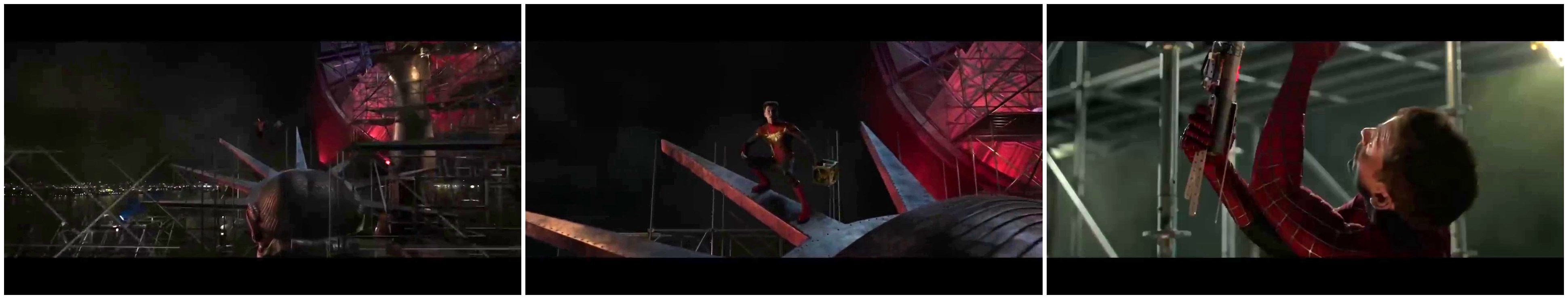}
  \caption{method: between local maxima, hanning window: 100}
  \label{gray_text}
\end{subfigure} 

\caption{Selected frames from the first 10 seconds of a sample video obtained using different experimental methods.}
\label{fig:extracted_frames}
\end{figure}

The motivation is that the first method, while more time efficient, targets exact moments of scene change, making it vulnerable to temporal attacks (e.g., insertion of random frames). In contrast, the second method may offer better resistance, yet requires more time to identify the frame. Using either method, we reduced the number of frames by 40 to over 150 times, which is 1.4 to 5.8 times more efficient than the standard 1~fps approach shown in~\cite{wang2023dual}. For a sample video with an fps of 24 and a total of 719 frames, we select 17 frames when using a Hanning window of size 30 (Fig.~\ref{fig:window30}), 13 frames with a Hanning window of size 50 (Fig.~\ref{fig:window50}), and only 5 frames with a Hanning window of size 100 (Fig.~\ref{fig:window100}). This corresponds to a reduction in the number of frames for a video by factors of approximately 42, 55, and 144, respectively. In contrast, a simple one-frame-per-second approach reduces the frames by a factor of only 24. However, it should be noted that smoothing with larger Hanning windows may result in the omission of frames from shorter scenes, as the larger the window, the greater the reduction in frames, which can be also observed in Fig.~\ref{fig:extracted_frames}, where we present selected frames from the first 10 seconds of the sample video obtained using different methods. This phenomenon is investigated during our experiments (cf.~Tab.~\ref{tab1}).  

\subsection{Resistance to Temporal Attacks\label{ssect:attack}}
To evaluate the robustness of both frame extraction methods, we conducted experiments with three types of temporal attacks:
\begin{description}
    \item[Random Frame Blackouts]:  The frames were blacked out in the original video with probabilities of $p = 1/25$ and $p = 1/10$. Naturally, the included black frames modify the related frames during video compression, affecting relative P- and B- frames, hence after video decoding more than 1 frame is affected. The number of affected frames depends on the blacked out frame, as it influences the I-, P-, and B-frame selection.
    \item[Targeted Frame Blackouts]: In this attack, the middle frame of each second of the video was blacked out in the same manner as in the first attack. Note that, such precise attacks are visually imperceptible to users, making them a significant threat to video copy detection systems.
    \item[Speed modification] We used ffmeg to modify the tempo at which the video is played. In technical terms, it keeps all the frames but saves them, as a video with changed fps, so that we get the effect of acceleration or deceleration. Naturally, it may influence the I-, P-, and B-frame selection.
\end{description}

\section{Experimental Results\label{sect:results}}
The winning team extracted a smaller validation set from the original training dataset for their experiments, reducing its size by a factor of four (1681 queries). We conduct our experiments using the same data.

\subsection{Parameter selection}
In order to analyze the efficacy of each frame selection method as well as the size of the smoothing window, we measured $\mu AP$ in a scenario reflecting the challenge. Moreover, we analyze the influence of the temporal attacks, described in Sect.~\ref{ssect:attack}. The results are shown in Tab.~\ref{tab1}.
One can note that the highest efficacy is obtained by the second selection strategy with Hanning window of size 30, with the same method and window size 50 being close second. Moreover, one can observe that the temporal attacks do have an impact on the matching efficacy, in particular Targeted attack influences the correct detection rate. As the first row of the table shows, an incorrectly selected frame representation can reduce the efficacy by up to 60\%.
Finally, one can note that the second selection strategy is more stable, resulting in significantly lower standard deviation, when compared with the same window size approach for the first strategy.

\begin{table}

\caption{Matching results [$\mu$AP] for proposed frames extraction methods depending on Hanning window size and applied temporal attack; \\ Local-max-windowX -- local maxima from the interframe differences curve, smoothed with Hanning window of size X, \\ Local-max-mid-windowX -- middle frames between local maxima from the interframe differences curve, smoothed with Hanning window of size X. The random attack is averaged in 3 independent runs, the standard deviation in given in parenthesis.\label{tab1}}
\begin{tabular}{|l|c|c|c|c|}
\hline
\diagbox{Method}{Attack type} & No attack & \makecell{Random attack \\ ($p = \frac{1}{25}$)} & \makecell{Random attack\\ ($p = \frac{1}{10}$)} & \makecell{Targeted attack\\ (Middle Frame/s)} \\
\hline
Local-max-window30 & \textbf{0.9300} & 0.8086 (0.0184) & 0.8012 (0.0167) & 0.3705 \\
Local-max-mid-window30 & \textbf{0.9300} & \textbf{0.8807} (0.0034) & \textbf{0.8747} (0.0021) & \textbf{0.8835} \\
Local-max-window50 & 0.9252 &  0.8361 (0.0068) & 0.8085 (0.0070) & 0.6891 \\
Local-max-mid-window50 & 0.9261 & 0.8549 (0.0013) & 0.8555 (0.0014) & 0.8577 \\
Local-max-window100 & 0.8273 & 0.7531 (0.0095) & 0.7564 (0.0106) & 0.8138 \\
Local-max-mid-window100 & 0.8343 & 0.7759 (0.0050) & 0.7667 (0.0016) & 0.8004 \\
\hline
\end{tabular}
\end{table}

\subsection{Efficiency\label{ssect:eff}}
Due to the ubiquitous nature of videos, the VCD system used in copyright management or DeepFake detection should be time efficient. Additionally, the more videos we want to track, the stored representation may be an issue, hence compact video representation is required.  As shown in Table~\ref{tab2}, our frame extraction method with smoothing window of size 50 reduces inference time by the factor of 2, allowing analysis of over 3 video samples per second, as compared to $1.58$ video sample per second in~\cite{wang2023dual}. Moreover, the representation of each video is reduced by almost 56\%. The relative speedup and representation size reduction are obtained with a minor performance trade-off, reducing the efficacy by less than 1 \% $\mu$AP. One can note, that reducing the size of the smoothing window results in higher efficacy (4\textperthousand{} increase in comparison to window size 50; 4\textperthousand{} worse than Dual-level), yet reduces significantly the time and memory gain. 

\begin{table}
\caption{Inference performance, total size of descriptors and matching results for original and proposed methods.\label{tab2}}
\begin{tabular}{|l|c|c|c|}
\hline
Method & \makecell{Inference \\ performance [vid/s]}  & \makecell{Total descriptors \\ size [Mb]} & \makecell{Match result [$\mu$AP]} \\
\hline
Dual-level~\cite{wang2023dual} &  1.58 & 57.09 & \textbf{0.9343} \\
Local-max-mid-window30 & 2.39 & 40.71 & 0.9300\\
Local-max-mid-window50 & \textbf{3.27} & \textbf{25.16} & 0.9261\\
\hline
\end{tabular}
\end{table}

\subsection{Robustness against temporal attacks}
Since the analysis in Sect~\ref{ssect:eff} includes tests on the dataset with image processing attacks, we want to investigate of the influence of the modifications typical for 3D data (a video in contrast to an image). For that purpose we modified the test dataset with attacks described in Sect.~\ref{ssect:attack}.

Table~\ref{tab3} presents the results under both Random and Targeted temporal attacks, showing the vulnerability of Dual-level detection. In particular, Targeted attack causes over 60\% drop in $\mu$AP with comparison to a scenario with no temporal attacks. On the other hand, our approach suffers around 5\% loss from the attakcs in case of window size 30, and 7\% in case of size 50. Those results show the robustness of the frame selection technique against temporal attacks, as well as the severity of the attack scope with only 1 frame per second being modified. Moreover, for random attacks, one can observe a lower standard deviation of the proposed method, indicating higher stability, with larger window resulting in lower variance. Note that the efficacy difference between two windows sizes is lower than 3\% for all the attack variants, showing the reasonability of using larger window, when the time and memory restrictions are severe.

Another typical temporal domain attack that is used to hinder copy detection is modification of video speed. The results of such tests are shown in Tab.~\ref{tab4}. We examined two factors of acceleration (1.2 and 1.5) and slowdown by the factor of 2. One can note, that the proposed method of frame extraction is invariant on the acceleration factor, whereas the Dual-level approach reduces its efficiency, the more, the faster the video is played. This is due to the deterministic frame selection which limits its adaptability to time compression or extension, resulting in almost 7\% drop, allowing our approach to outperform it by similar degree.

\begin{table}
\caption{Matching results [$\mu$AP] for dual-level and proposed methods depending on applied temporal attack: Frame blackouts. For random type attacks the result is averaged over 3 attempts, the standard deviation in given in parenthesis.\label{tab3}}
\begin{tabular}{|l|c|c|c|c|}
\hline
\diagbox{Method}{Attack type} & No attack & \makecell{Random attack \\ ($p = \frac{1}{25}$)} & \makecell{Random attack\\ ($p = \frac{1}{10}$)} & \makecell{Targeted attack\\ (Middle Frame/s)} \\
\hline
Dual-level~\cite{wang2023dual} & \textbf{0.9343} & 0.8788 (0.0066) & 0.8674 (0.0072) & 0.3705 \\
Local-max-mid-window30 & 0.9300 & \textbf{0.8807} (0.0034) & \textbf{0.8747} (0.0021) & \textbf{0.8835} \\
Local-max-mid-window50 & 0.9261 & 0.8549 (0.0013) & 0.8555 (0.0014)  & 0.8577 \\
\hline
\end{tabular}
\end{table}

\begin{table}
\caption{Matching results [$\mu$AP] for dual-level and proposed methods depending on a dataset with Speed Modification attack.\label{tab4}}
\begin{tabular}{|l|c|c|c|c|}
\hline
\diagbox{Method}{Acceleration factor} & 1.0 (No attack) & 1.2 & 1.5 & 0.5 \\
\hline
Dual-level~\cite{wang2023dual} & \textbf{0.9343} & 0.9052 & 0.8709 & 0.9284 \\
Local-max-mid-window30 & 0.9300 & \textbf{0.9300} & \textbf{0.9300} & \textbf{0.9300} \\
Local-max-mid-window50 & 0.9261 & 0.9261 & 0.9261  & 0.9261 \\
\hline
\end{tabular}
\end{table}


\section{Conclusion}

This paper focuses on the problem of Video Copy Detection and addresses the limitations of the Dual-level detection method~\cite{wang2023dual} that was successful in Meta AI Challange. We propose an improved method of video representation by more adaptive frame selection. Additionally, we analyze the performance of the detection method against three proposed temporal attacks. Using local maxima of interframe differences, the proposed method reduces computational costs while keeping comparable efficiency measured as micro-average precision. The performance difference on the DVSC2023 dataset is less than 1\%, while increasing resilience against temporal attacks. Our experiments show that our method reduces the number of frames required to properly represent a video by  1.4 to 5.8 times when compared to the standard one-frame-per-second approach. Moreover, in the performed tests, our method achieved 2 times faster inference, which is important in real-world applications when processing massive video databases. Moreover, the proposed approach efficacy showed to be invariant on video speed manipulations, whereas the previous method suffered a 7\% $\mu$AP drop. Similar results were obtained for frame blackout, both random and targeted, showing the resistance of our method.

Future research will focus on developing adaptive temporal alignment techniques to enhance robustness against significant frame modifications (like blackouts), as well as integrating feature matching techniques that allow reducing the size of representation for even better performance enhancement. Additionally, expanding the approach to handle localized transformations and cross-modal attacks (e.g., temporal modifications, room attack and overlays as shown in Fig.~\ref{fig:transformations}) will further strengthen its effectiveness. Another branch of research is further improvement of inference time, so that large-scale video databases may be checked for copies. Finally, the VCD models should be investigated for their interpretability, so that they follow XAI research trend and be compliant with jurisdical restrictions for real-world legal usage.

\begin{credits}
\subsubsection{\ackname} This work has been partially funded by Department of Artificial Intelligence, Wroclaw University of Science and Technology.

\subsubsection{\discintname}
 The authors have no competing interests to declare that are
relevant to the content of this article.
\end{credits}
%
%
%
\bibliographystyle{splncs04}
\bibliography{references}

\begin{thebibliography}{10}
\providecommand{\url}[1]{\texttt{#1}}
\providecommand{\urlprefix}{URL }
\providecommand{\doi}[1]{https://doi.org/#1}

\bibitem{arnab2021vivit}
Arnab, A., Dehghani, M., Heigold, G., Sun, C., Lu{\v {c}}i{\'c}, M., Schmid, C.: Vivit: A video vision transformer. In: Proceedings of the IEEE/CVF international conference on computer vision. pp. 6836--6846 (2021)

\bibitem{black2023vader}
Black, A., Jenni, S., Bui, T., Tanjim, M.M., Petrangeli, S., Sinha, R., Swaminathan, V., Collomosse, J.: Vader: Video alignment differencing and retrieval. In: Proceedings of the IEEE/CVF International Conference on Computer Vision. pp. 22357--22367 (2023)

\bibitem{chiang2022multi}
Chiang, T.H., Tseng, Y.C., Tseng, Y.C.: A multi-embedding neural model for incident video retrieval. Pattern Recognition  \textbf{130},  108807 (2022)

\bibitem{deng20233d}
Deng, R., Wu, Q., Li, Y.: 3d-csl: self-supervised 3d context similarity learning for near-duplicate video retrieval. In: 2023 IEEE International Conference on Image Processing (ICIP). pp. 2880--2884. IEEE (2023)

\bibitem{deng2024differentiable}
Deng, R., Wu, Q., Li, Y., Fu, H.: Differentiable resolution compression and alignment for efficient video classification and retrieval. In: ICASSP 2024-2024 IEEE International Conference on Acoustics, Speech and Signal Processing (ICASSP). pp. 3200--3204. IEEE (2024)

\bibitem{fojcik2025extremely}
Fojcik, K., Syga, P., Klonowski, M.: Extremely compact video representation for efficient near-duplicates detection. Pattern Recognition  \textbf{158},  111016 (2025)

\bibitem{he2016deep}
He, K., Zhang, X., Ren, S., Sun, J.: Deep residual learning for image recognition. In: Proceedings of the IEEE conference on computer vision and pattern recognition. pp. 770--778 (2016)

\bibitem{kim2024}
Kim, J., Woo, S., Nang, J.: Relational self-supervised distillation with compact descriptors for image copy detection (2024), \url{https://arxiv.org/abs/2405.17928}

\bibitem{kordopatis2019visil}
Kordopatis-Zilos, G., Papadopoulos, S., Patras, I., Kompatsiaris, I.: Visil: Fine-grained spatio-temporal video similarity learning. In: Proceedings of the IEEE/CVF international conference on computer vision. pp. 6351--6360 (2019)

\bibitem{kordopatis2017near2}
Kordopatis-Zilos, G., Papadopoulos, S., Patras, I., Kompatsiaris, Y.: Near-duplicate video retrieval with deep metric learning. In: Proceedings of the IEEE international conference on computer vision workshops. pp. 347--356 (2017)

\bibitem{li2020two}
Li, J., Zhang, H., Wan, W., Sun, J.: Two-class 3d-cnn classifiers combination for video copy detection. Multimedia Tools and Applications  \textbf{79}(7-8),  4749--4761 (2020)

\bibitem{ma2024let}
Ma, Z., Dong, J., Ji, S., Liu, Z., Zhang, X., Wang, Z., He, S., Qian, F., Zhang, X., Yang, L.: Let all be whitened: Multi-teacher distillation for efficient visual retrieval. In: Proceedings of the AAAI Conference on Artificial Intelligence. vol.~38, pp. 4126--4135 (2024)

\bibitem{pizzi20242023}
Pizzi, E., Kordopatis-Zilos, G., Patel, H., Postelnicu, G., Ravindra, S.N., Gupta, A., Papadopoulos, S., Tolias, G., Douze, M.: The 2023 video similarity dataset and challenge. Computer Vision and Image Understanding  \textbf{243},  103997 (2024)

\bibitem{pizzi2022self}
Pizzi, E., Roy, S.D., Ravindra, S.N., Goyal, P., Douze, M.: A self-supervised descriptor for image copy detection. Proc. CVPR  (2022)

\bibitem{simonyan2014very}
Simonyan, K., Zisserman, A.: Very deep convolutional networks for large-scale image recognition. In: Bengio, Y., LeCun, Y. (eds.) 3rd International Conference on Learning Representations, {ICLR} 2015, San Diego, CA, USA, May 7-9, 2015, Conference Track Proceedings (2015)

\bibitem{wang2023dual}
Wang, T., Ma, F., Liu, Z., Rao, F.: A dual-level detection method for video copy detection. arXiv preprint arXiv:2305.12361  (2023)

\bibitem{wu2007practical}
Wu, X., Hauptmann, A.G., Ngo, C.W.: Practical elimination of near-duplicates from web video search. In: Proceedings of the 15th ACM international conference on Multimedia. pp. 218--227 (2007)

\bibitem{ZHONG2024103863}
Zhong, J.L., Gan, Y.F., Yang, J.X.: Efficient detection of intra/inter-frame video copy-move forgery: A hierarchical coarse-to-fine method. Journal of Information Security and Applications  \textbf{85},  103863 (2024). \doi{https://doi.org/10.1016/j.jisa.2024.103863}, \url{https://www.sciencedirect.com/science/article/pii/S2214212624001650}

\end{thebibliography}
%




\end{document}